\documentclass[letterpaper,UKenglish,11pt]{article}

\usepackage{authblk}
\usepackage[top=2.5cm, bottom=2.5cm, left=2.5cm, right=2.5cm]{geometry}

\usepackage[utf8]{inputenc} 
\usepackage[T1]{fontenc}    
\usepackage{hyperref}       
\usepackage{url}            
\usepackage{booktabs}       
\usepackage{amsfonts}       
\usepackage{nicefrac}       
\usepackage{microtype}      
\usepackage{xcolor}
\usepackage{soul}
\usepackage{wrapfig}
\usepackage{booktabs}
\usepackage{multirow}
\usepackage{mathtools}
\usepackage{graphicx}
\usepackage{float}
\usepackage[caption = false]{subfig}
\usepackage{xspace}
\usepackage{xparse}
\usepackage{caption}
\usepackage[textsize=footnotesize]{todonotes}

\newcommand{\prior}{{p_{\theta}}}
\newcommand{\post}{{q_{\phi}}}
\newcommand{\eye}{\mathbf{I}}

\newcommand{\EE}{\mathbb{E}}

\DeclarePairedDelimiterX{\infdivx}[2]{(}{)}{%
  #1\;\delimsize|\delimsize|\;#2%
}
\newcommand{\dkl}[2]{\ensuremath{\mathrm{KL}\infdivx{#1}{#2}}\xspace}

\title{WeLa-VAE: Learning Alternative Disentangled Representations Using Weak Labels}

\author{Vasilis Margonis
\qquad Athanasios Davvetas \qquad Iraklis A. Klampanos \\
\small
Institute of Informatics and Telecommunications\\
National Center for Scientific Research ``Demokritos''\\
Agia Paraskevi 15341, Athens, Greece \\
\texttt{\{vmargonis, tdavvetas, iaklampanos\}@iit.demokritos.gr}}

\date{}

\begin{document}

\maketitle

\begin{abstract}
Learning disentangled representations without supervision or inductive biases, often leads to non-interpretable or undesirable representations.
On the other hand, strict supervision requires detailed knowledge of the true generative factors, which is not always possible. 
In this paper, we consider weak supervision by means of high-level labels that are not assumed to be explicitly related to the ground truth factors.
Such labels, while being easier to acquire, can also be used as inductive biases for algorithms to learn more interpretable or alternative disentangled representations.
To this end, we propose WeLa-VAE, a variational inference framework where observations and labels share the same latent variables, which involves the maximization of a modified variational lower bound and total correlation regularization.
Our method is a generalization of TCVAE, adding only one extra hyperparameter.
We experiment on a dataset generated by Cartesian coordinates and we show that, while a TCVAE learns a factorized Cartesian representation, given weak labels of distance and angle, WeLa-VAE is able to learn and disentangle a polar representation.
This is achieved without the need of refined labels or having to adjust the number of layers, the optimization parameters, or the total correlation hyperparameter.
\end{abstract}


\section{Introduction}
In representation learning, it is often assumed that complex, high-dimensional data, like face photos, are generated by a small number of mutually independent latent variables, usually referred to as generative factors.
This assumption implies that such data can be explained by simple explanatory features which are significantly fewer than the original dimensions.
Learning representations which identify and separate the few distinct, informative factors of variation is termed as disentanglement.
Although a formal definition is currently under debate, a representation is vaguely characterized as disentangled if single features are sensitive to changes in single generative factors, while being relatively invariant to changes in other factors \cite{Bengio2013}.
It has been suggested that disentangled representations, despite their potential on generalizing to diverse downstream tasks,  can lead to better understanding of a dataset's underlying distribution, provide intuition or allow for generative tasks that require ``perception'', such as conditional generation \cite{Bengio2013, LakeUTG2017}.

Locatello et al. \cite{LocatelloBLRGSB19} proved that disentanglement is impossible without inductive biases on both the model and the data, as there exist datasets with multiple sets of generative factors.
In the same paper, large-scale experimentation indicated that current unsupervised models cannot reliably learn disentangled representations as the choice of random seeds and hyperparameters seem to have a greater impact than the choice of the model itself.
Moreover, the assumption that a disentangled representation is useful for downstream tasks could not be validated for the considered models and datasets.
The authors concluded that future work on disentanglement should deviate from the purely unsupervised setting by making the role of inductive biases and supervision more explicit.

On the other hand, introducing strict supervision requires prior knowledge on the nature or number of the true generative factors. 
However, this information is often either unavailable or expensive, if not impossible to acquire. 
For instance, there may exist practical examples of datasets for which no generative factors stand out enough to be recognized upon inspection.
In addition, labelling can be laborious and usually targets specific downstream tasks or domain applications, not focusing on the implicit properties of the data.
There may be cases where labels reflect generative factors, e.g. digit labels of the MNIST dataset, but this is not generally true in realistic datasets.

An approach that allows the inclusion of supervision while mitigating the difficulties of acquiring ground truth labels is weak supervision.
Weak supervision is often provided via weak labels, which may be noisy (e.g. produced by non-experts), partially available (also known as semi-supervision), high-level (dividing the data into fewer classes), or non-corresponding to ground truth.
In this paper, we consider the exploitation of weak labels of the last two categories in the context of disentanglement.
This information can act as inductive bias for learning disentangled representations,
without posing the restrictions of strict supervision or the pitfalls of unsupervised practices.
Moreover, weak supervision may lead to representations that are impossible for unsupervised models to learn, hence more interpretable and potentially useful for downstream tasks.

In this paper we contribute:
\begin{enumerate}
    \item WeLa-VAE, a scalable framework based on Variational Auto-encoders \cite{KingmaW14} with \emph{total correlation} regularization, that leverages weak labels towards learning disentangled representations.
    \item A synthetic dataset of images depicting white Gaussian blobs on a black canvas, along with appropriate weak labels to support our experiments.
    \item A thorough quantitative and qualitative evaluation, and comparison of the models used for the considered tasks based on a suitable novel and generalizable metric.
\end{enumerate}

\begin{figure}[ht]
    \centering
    \subfloat[Unsupervised $\beta$-TCVAE.]{
        \includegraphics[width=.48\linewidth]{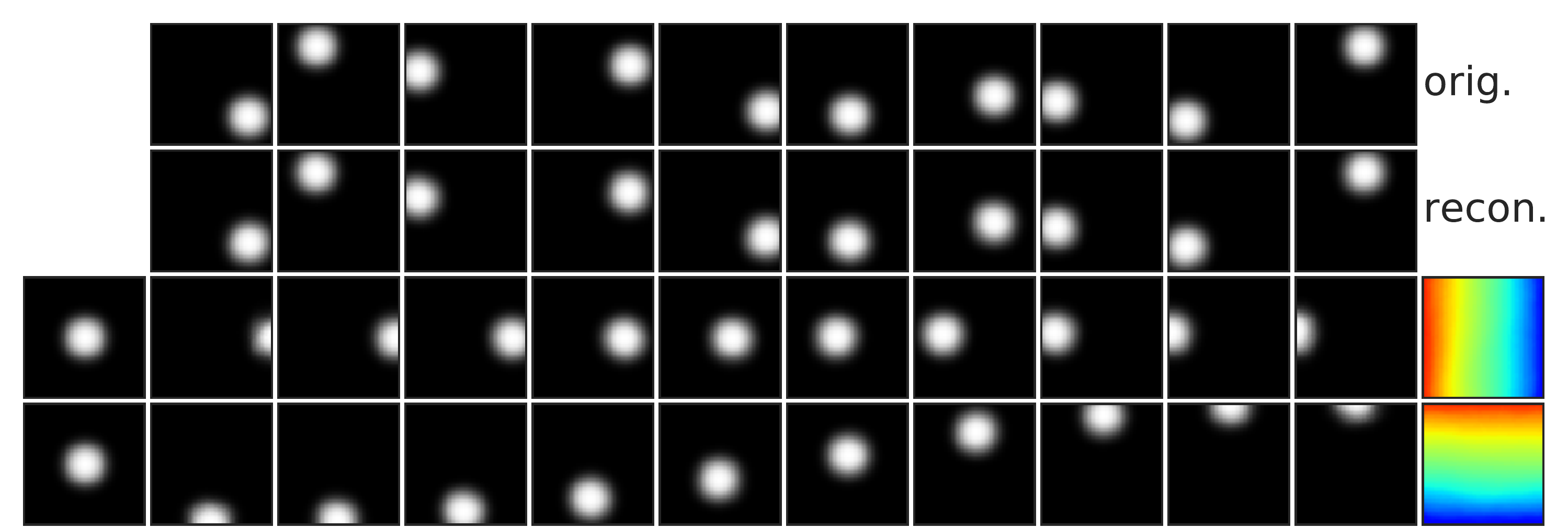}
        \label{fig:introtcvae}
    } 
    \subfloat[WeLa-VAE with polar subdivision labels.]{
        \includegraphics[width=.48\linewidth]{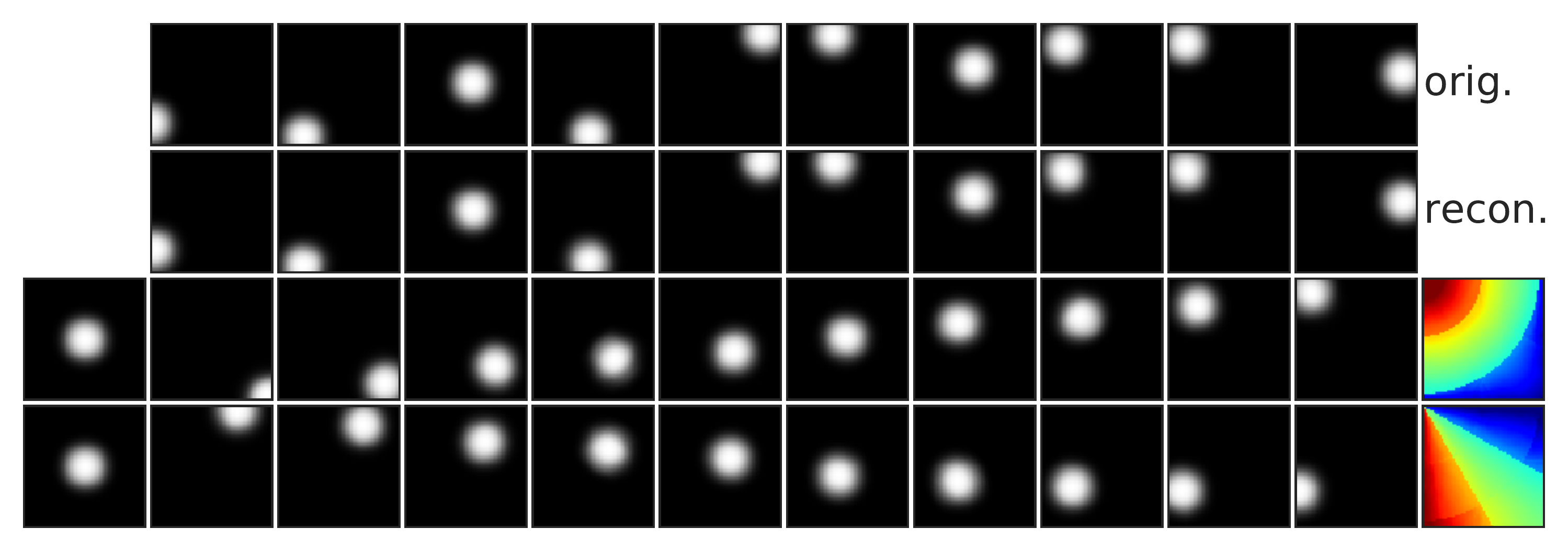}
        \label{fig:introwelavae}
    }
    \caption{$\beta$-TCVAE and WeLa-VAE trained on Blobs dataset. First row: original images. Second row: corresponding reconstructions. Rest: latent traversals in $[-3,+3]$. Left-most column: Image before traversal. Heat maps show the mean activation of each channel, as a function of position (dark blue, green, and dark red correspond to $-3$, $0$ and $+3$, respectively).}
    \label{fig:introcomparison}
\end{figure}

Next section discusses related work. 
In section \ref{section:derivation} we describe the derivation and implementation of WeLa-VAE. 
In Section \ref{section:experiments}, we introduce the dataset, provide details about the evaluation method and discuss the performance of WeLa-VAE next to a baseline established by $\beta$-TCVAE. 
Finally, we highlight the strengths and weaknesses of our framework, and propose directions for future work in section \ref{section:conclusion}.

\section{Related Work} \label{section:relatedwork}

The literature on learning disentangled representations can be divided in unsupervised and supervised learning frameworks. 
Unsupervised models are trained without available information about the number or the nature of the generative factors. 
Notable earlier approaches based on spike-and-slab restricted Boltzmann machines \cite{Desjardins2012}, tensor analyzers \cite{Tang2013} and commutative Lie groups \cite{Cohen2014}, produced promising results, albeit failing to scale well on big datasets. 
InfoGAN \cite{Chen2016} although scalable, inherits training instability and mode jumping from GAN \cite{Goodfellow2014}.

Kingma and Welling \cite{KingmaW14} introduced variational auto-encoders (VAEs) as a scalable framework for variational Bayesian inference.
In VAE, a two-step generative process is assumed where a latent vector $z$ is sampled from an isotropic Gaussian prior $p(z)=\mathcal{N}(0, \eye)$ and observations $x$ are sampled from the posterior $\prior(x|z)$. 
The distribution $\prior(z|x)$ is approximated by a variational distribution $\post(z|x)$. 
Both $\prior(x|z)$ and $\post(z|x)$ are parameterized by neural-networks, which are jointly trained by maximizing the variational lower bound (ELBO) of $\log{\prior(x)}$ for all observations $x$:
\begin{equation} \label{vaeobj}
   \mathrm{ELBO}(\theta, \phi) =  -\dkl{\post(z|x)}{p(z)} + \EE_{z \sim \post}[\log{\prior(x|z)}]. 
\end{equation}

State-of-the-art unsupervised methods for disentanglement are variants of VAE which enforce a factorized aggregated posterior $\post(z) = \int \post(z|x) \prior(x) dx$. 
In $\beta$-VAE \cite{Higgins2017} a Lagrange multiplier on the KL term introduces an information bottleneck on the latent channel with a trade-off in reconstruction quality. 
Burgess et al. \cite{Burgess2018} proposed to incrementally increase the latent channel capacity for better reconstructions.
Independently, FactorVAE \cite{Kim2018} and $\beta$-TCVAE \cite{Chen2018} augment the VAE objective \eqref{vaeobj} with a regularizer that penalizes the \emph{total correlation} $\dkl{q(z)}{\prod_k q(z_k)}$.
DIP-VAE \cite{Kumar2018} uses a similar augmentation which enforces the moments of $q(z)$ and the isotropic Gaussian prior $p(z)$ to match.

Locatello et al. \cite{LocatelloBLRGSB19} performed a large-scale empirical study on the unsupervised setting involving VAE variants \cite{Higgins2017, Burgess2018, Kim2018, Chen2018, Kumar2018}, disentanglement metrics \cite{Higgins2017, Kim2018, Eastwood2018, Kumar2018, Chen2018, Ridgeway2018} and synthetic datasets.
Their experimental results indicate that the choice of hyperparameters and random seeds is more significant than model selection on the learned representations and metric scores, and that good random seeds and hyperparameters cannot be identified without access to ground-truth labels.
They suggest that unsupervised learning of disentangled representation is unreliable and motivate future work that focuses on inductive biases and explicit supervision.
WeLa-VAE framework is a step towards that direction.

In the supervised scheme, factors of interest are explicitly labelled, and information about their nature (e.g. categorical or continuous) is available.
However, obtaining explicit labels is problematic, therefore most models are either semi or weakly supervised. For example,  Kingma et al. \cite{KingmaMRW14}, Narayanaswamy et al. \cite{NarayanaswamyPM17} and Yan et al. \cite{YanYSL16} propose semi-supervised variants of VAEs where observed labels are treated as latent variables. 
Other approaches considered various forms of weak supervision, i.e., implicit information about the factors of variation, including temporal coherence \cite{DentonB17, HsuZG17, VillegasYHLL17}, rendering knowledge in computer graphics \cite{Kulkarni2015}, and leveraging groups of observations where the generative factors are constant \cite{BouchacourtTN18, FengWKZTS18, RuizMBV2019}. 
In contrast to existing supervised approaches, we provide weak supervision by high-level labels which are neither assumed to be explicitly related to the ground truth factors, nor are they treated are latent variables.

\section{WeLa-VAE Framework} \label{section:derivation}

\begin{figure}[ht]
\centering
\includegraphics[width=0.15\linewidth]{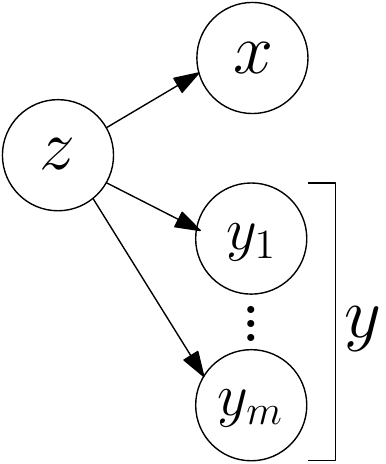}
\caption{Generative model}
\label{fig:genmodel}
\end{figure}

Let $\mathcal{D}=\{X, \mathcal{Y}=(Y_1, \ldots, Y_m)\}$ be a dataset consisting of observations $x \in \mathbb{R}^D$ and $m$ weak labels $y_1, \ldots, y_m$ per observation. 
That is, for every $x^i$ we have a corresponding multi-label $y^i=(y^i_1, \ldots, y^i_m)$.
Given such a dataset, our objective is to obtain a disentangled representation of $x$ that is influenced by $y$.
Given that the labels are related to the observations, it is reasonable to consider the case where $y$ shares the same latent variables with $x$.
Therefore, we assume an extended generative model depicted in Figure \ref{fig:genmodel}, which is similar to that of VAE \cite{KingmaW14}, with the addition of one variable for each label. 
More specifically, a latent variable is first sampled from a distribution $p(z)$. Then, $x$ and $y=(y_1, \ldots, y_m)$ are sampled from the conditionals $\prior(x|z)$ and $\prior(y|z) = \prod_{j} \prior(y_j|z)$, respectively.

A suitable objective function could be the maximization of the log-likelihood of $\prior(x, y)$.
Using an auxiliary distribution $\post(z|x,y)$ to approximate the intractable posterior $\prior(z|x,y)$, allows us to obtain a tractable variational lower bound\footnote{See Appendix \ref{appendixA:elbo}} on the log-probability:
\begin{equation} \label{equation:elbo}
    \begin{aligned}
    \log{\prior(x,y)} &\geq  \EE_{z \sim \post}[\log{\prior(x|z)}] + \EE_{z \sim \post}[\log{\prior(y|z)}] -\dkl{\post(z|x,y)}{p(z)} \\
    &= \EE_{z \sim \post}[\log{\prior(x|z)}] + \sum_{j=1}^m \EE_{z \sim \post}[\log{\prior(y_j|z)}] -\dkl{\post(z|x,y)}{p(z)}. 
\end{aligned}
\end{equation}
However, we are not interested in the exact value of the log-probability, as our objective is to obtain disentangled representations that are influenced by the weak-labels.
Therefore, we choose to optimize a modified variational lower bound.
More specifically, to ensure that the labels are reconstructed properly, we add a multiplier hyperparameter $\gamma \geq 1$ on the labels' reconstruction term to control its relative scale to the rest of the terms:
\begin{equation} \label{equation:modifiedelbo}
    \gamma\mathrm{ELBO}(\theta, \phi) := \EE_{z \sim \post}[\log{\prior(x|z)}] + \gamma \cdot \sum_{j=1}^m \EE_{z \sim \post}[\log{\prior(y_j|z)}] - \dkl{\post(z|x,y)}{p(z)}.
\end{equation}
Moreover, as in \cite{Kim2018, Chen2018}, to enforce a factorized aggregated posterior
\begin{equation*}
    \post(z) = \iint \post(z|x,y) \cdot \prior(x) \cdot \prior(y) \ \mathrm{d} x \mathrm{d} y,
\end{equation*}
we introduce a total correlation regularizer controlled by a hyperparameter $\beta \geq 0$.
Since this term is intractable, we use the biased Monte-Carlo estimator proposed in $\beta$-TCVAE \cite{Chen2018}. 
Note that WeLa-VAE generalizes easily to different ways of forcing a factorized aggregated posterior. For example, one may substitute the total correlation penalty with the regularizers of DIP-VAE \cite{Kumar2018}.
The objective function of WeLa-VAE is
\begin{equation} \label{equation:WeLa-VAEloss}
\min\mathcal{L}(\theta, \phi) = -\EE_{x, y} [\gamma\mathrm{ELBO}(\theta, \phi)] + \beta \cdot \dkl{\post(z)}{\prod_{k=1}^K \post(z_k)}.
\end{equation}

We postulate that this method results in representations which are strongly affected by the labels, as the network is forced to gain the discriminative ability to reconstruct the labels accurately.
Meanwhile, the total correlation penalty encourages the representation to factorize.
As in most variants of VAE's, the distributions $\post(z|x,y)$, $\prior(x|z)$ and $\prior(y|z)$ are parameterized by two neural networks, one for $\post(z|x,y)$ (encoder) and one for both $\prior(x|z)$ and $\prior(y|z)$ (decoder).
The prior $p(z)$ is set to the non-parametric isotropic Gaussian $\mathcal{N}(0, \eye)$ and $\post(z|x,y)$ is also a diagonal Gaussian $\mathcal{N}(\mu, \sigma^2 \eye)$. 
Sampling from $\post(z|x,y)$ is done via the reparameterization trick, i.e. $z {=} \mu {+} \sigma {\odot} \epsilon$, where $\epsilon \sim \mathcal{N}(0, \eye)$. 
The choice of family of distributions for $\prior(x|z)$ and $\prior(y|z)$ depends on the nature of the dataset.
During our implementation of WeLa-VAE, $x$ and $y$ are concatenated as a single input in a stacked dense autoencoder. However, the implementation of WeLa-VAE can be versatile, allowing for the utilisation of the intrinsic properties of any given dataset. For instance, WeLa-VAE can be adapted to implementations that utilise the spatial aspect of a dataset, such as convolutional layers, by concatenating weak labels along with the first dense layer representations of $x$.

\section{Experiments} \label{section:experiments}
In this section, we introduce a dataset for experimentation, provide details about the experimental parameters and evaluation method and discuss the performance of WeLa-VAE compared to a baseline established by $\beta$-TCVAE.

\subsection{Evaluation Dataset}
\begin{figure}[ht]
    \centering
    \subfloat[Blob sample]{
    \includegraphics[width=.17\linewidth]{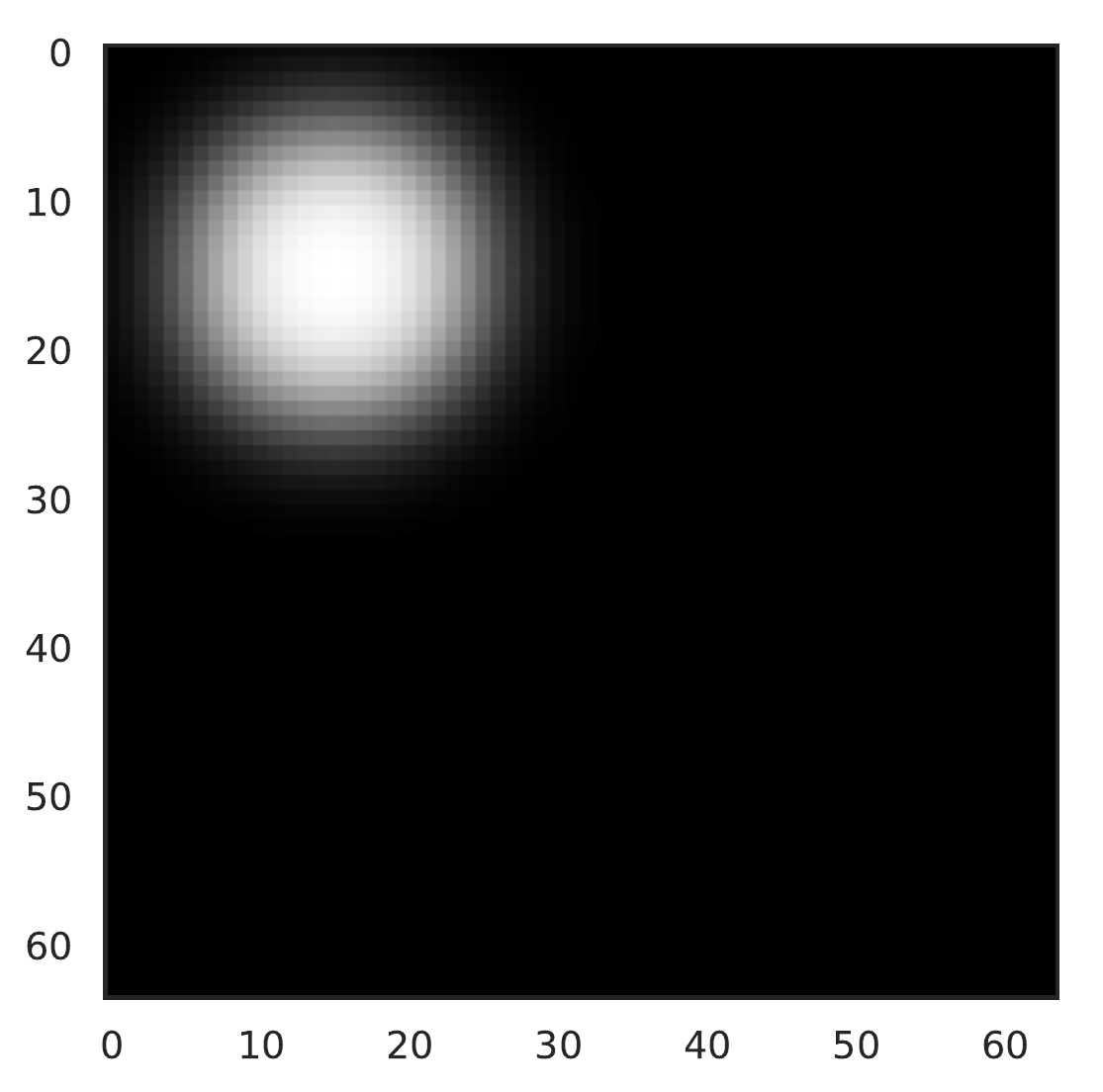}
    \label{fig:blobsample}
    }
    \quad
    \subfloat[Division of the $64{\times}64$ canvas into disjoint regions for membership labels. The number of regions determines the dimension of the resulting one-hot encoded vector of the label. Figures are presented in ascending order of label dimension ($p$). Top Row: angle; bottom Row: distance.]{
    \includegraphics[width=.77\linewidth]{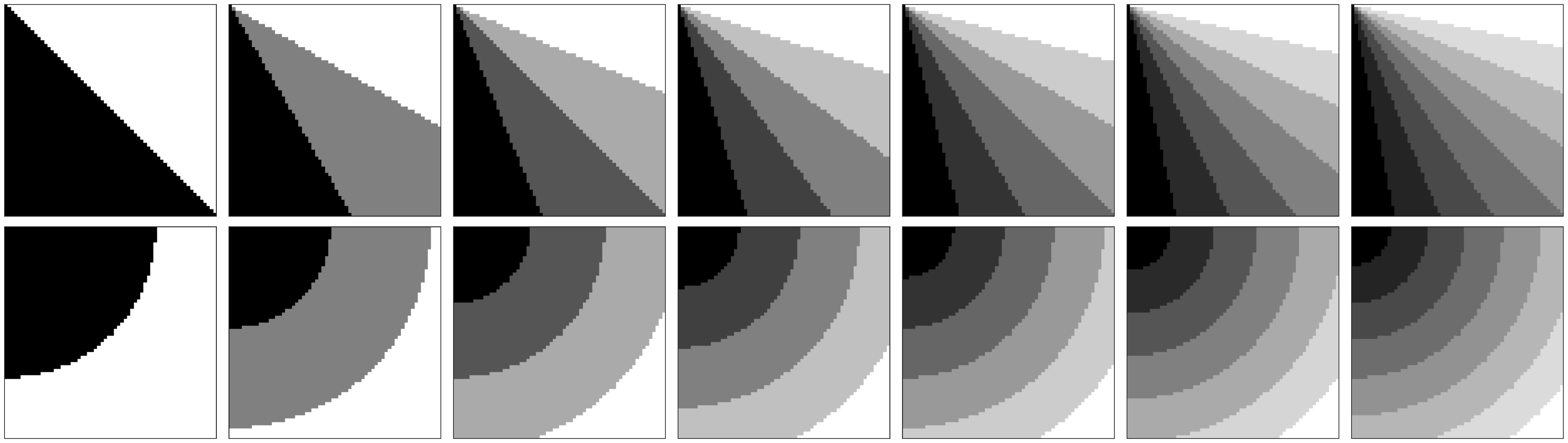}
    \label{fig:weaklabels}
    }
    \caption{Blobs dataset sample and division of canvas for weak labels.}
    \label{fig:dataset}
\end{figure}

We test WeLa-VAE on a synthetic dataset of $64{\times}64$ images containing a white Gaussian blob positioned on a black canvas (Figure \ref{fig:blobsample}).
A similar dataset has appeared in the context of disentanglement in \cite{Burgess2018}. 
In our case, the use of weak labels and the evaluation methodology necessitated the creation of a new dataset.
The generative factors are the Cartesian coordinates of the blob's center on the canvas. 
For every possible position, we sample 25 different Gaussian blobs, getting a total of $N=102,400$ images, which is the training set $X$.
This is a motivating example because the observed images can be also explained in polar coordinates, therefore its a case of a dataset which has two different, but equally valid sets of generative factors.
Our objective is to provide WeLa-VAE with weak labels indicating angle and distance, expecting to obtain a disentangled polar representation.
Assuming that $(0,0)$ lies at the top left corner of the canvas, we construct two membership labels, one for angle and one for distance, for every image (Figure \ref{fig:weaklabels}).
To this end, the canvas is divided in $p$ disjoint areas, and the label vector is the one-hot encoding of the label associated with the area that the center of the blob lies in.
The value $p$ determines the dimension of the one-hot encoded label vectors, and it is very important in this context as it empirically indicates the amount of bias introduced by the labels: 
Higher $p$ means that more decision boundaries must be learned by the network to reconstruct the labels accurately.
Hence, to test the effect of label dimensionality on the learned representations, we construct labels for $p=2,3,\ldots,8$.

\subsection{Evaluation}
We consider two tasks: (1) learning a factorized Cartesian representation and (2) learning a factorized polar representation.
We use latent traversals and positional heat maps as a means of qualitative evaluation.
Quantitative evaluation of WeLa-VAE is not trivial and known disentanglement metrics are not suitable in this case.
Specifically, such metrics measure various notions of statistical relation between the learned representation and the ground truth factors, which is unfit for the second task of learning a polar representation.
Thus, we choose to measure performance as the ability of approximating the true Cartesian coordinates $(c_1, c_2)$ used to generate an image $x$, by a simple transformation of the learned latent space.
Note that we use the mean vector $\mu$ of the Gaussian encoder as the representation, not the sample $z$.

For the task of learning a factorized Cartesian representation, an ideal model learns a representation $\mu$ with one channel $\mu_i$ corresponding to $c_1$ and one $\mu_j$ corresponding to $c_2$. An approximation $(\tilde{c_1}, \tilde{c_2})$ can be computed by simply mapping the feature value ranges to $[0, 64]$ linearly:
\begin{equation} \label{eq:tcvaemetric}
\begin{aligned}
\tilde{c_1} &= (\mu_i - \min(\mu_i)) \cdot (64 / (\max(\mu_i)-\min(\mu_i)), \\
\tilde{c_2} &= (\mu_j - \min(\mu_j)) \cdot (64 / (\max(\mu_j)-\min(\mu_j)).
\end{aligned}
\end{equation}

For the second task, we approximate $(c_1, c_2)$ as in a similar fashion:
Let $\mu_i$ correspond to angle $\phi$ and $\mu_j$ correspond to distance $d$. First, we compute
\begin{equation} \label{eq:welametric1}
\begin{aligned}
    \mu_{\phi} &= (\mu_i - \min(\mu_i)) \cdot ((\pi/2)/(\max(\mu_i)-\min(\mu_i)), \\
    \mu_d &= (\mu_j - \min(\mu_j)) \cdot (90.5 / (\max(\mu_j)-\min(\mu_j)),
\end{aligned} 
\end{equation}
to ensure that the values lie in a valid range.
Since $c_1, c_2 \in [0, 64]$, then $d \in [0, 90.5]$ and $\phi \in [0, \pi/2]$. 
Then, we apply the known mapping from polar to Cartesian coordinates,
\begin{equation} \label{eq:welametric2}
\begin{aligned}
    \tilde{c_1} &= \mu_d \cdot \cos(\mu_{\phi}), \\
    \tilde{c_2} &= \mu_d \cdot \sin(\mu_{\phi}).
\end{aligned}
\end{equation}

We measure approximation as the mean squared $L2$ error of $(c_1, c_2)$ from $(\tilde{c_1}, \tilde{c_2})$, across all samples. 
However, there is no way of knowing which channels to assign to angle and distance without qualitative inspection. 
Moreover, it is possible for the encoder to invert the order of values, e.g. blobs in distance close to $0$ represented by high positive values instead of low negative, and vice versa. 
Therefore, we try all possible assignments and inversions and return the lowest MSE.
For the task of learning a Cartesian representation, we take approximations via equations \eqref{eq:tcvaemetric} and measure MSE, while for the task of learning a polar representation, we use equations \eqref{eq:welametric1} and \eqref{eq:welametric2}.
This evaluation protocol easily generalizes to other tasks where certain representation qualities are expected.

\subsection{Experimental setup}
Images are valued in $[0,1]$, therefore we parameterize the distribution $\prior(x|z)$ as a multivariate Bernoulli. One-hot encoded labels are binary vectors thus $\prior(y|z)$ is parameterized as categorical.
In all experiments, both encoder and decoder consist of two fully-connected, $1200$--neuron layers. 
Training is done using the Adam optimizer \cite{KingmaB14} with learning rate $10^{-4}$, in batches of size 256 for 150 epochs\footnote{Each epoch takes approximately 9 seconds on a GeForce 1080Ti.}. 
We set $\beta{=}40$ for both TCVAE and WeLa-VAE to showcase that good $\beta$ values can be transferred.
As we measure reconstruction error with cross-entropy for both images and labels, we choose $\gamma$ such that 
\begin{equation*}
\textrm{Image Dimension} \approx \gamma \times \textrm{Label Dimension},
\end{equation*}
to ensure that image and label reconstructions are of similar scale.
Although this rule-of-thumb proved sufficient for the concerned dataset and task, searching for $\gamma$ can also be done through label reconstruction loss, e.g., choosing a value $\gamma$ that results in accuracy greater than some specified threshold.
Eight models are tested; one TCVAE with latent channel size $K{=}5$, and one WeLa-VAE with $K{=}2$ for every label dimensionality $p$\footnote{Choosing $K>m$ produced sub-optimal results, as redundant channels were opened, leading to entangled representations. Allowing for arbitrarily large $K$ is left as future work.}.
We train each model for 50 different types of random weight initialization and report the scores for each task.
We also visualize the representations that achieve the lowest MSEs.

\subsection{Results}
\begin{figure}[ht]
    \centering
    \subfloat[Lowest MSE ($9.24$)]{
    \includegraphics[width=.48\linewidth]{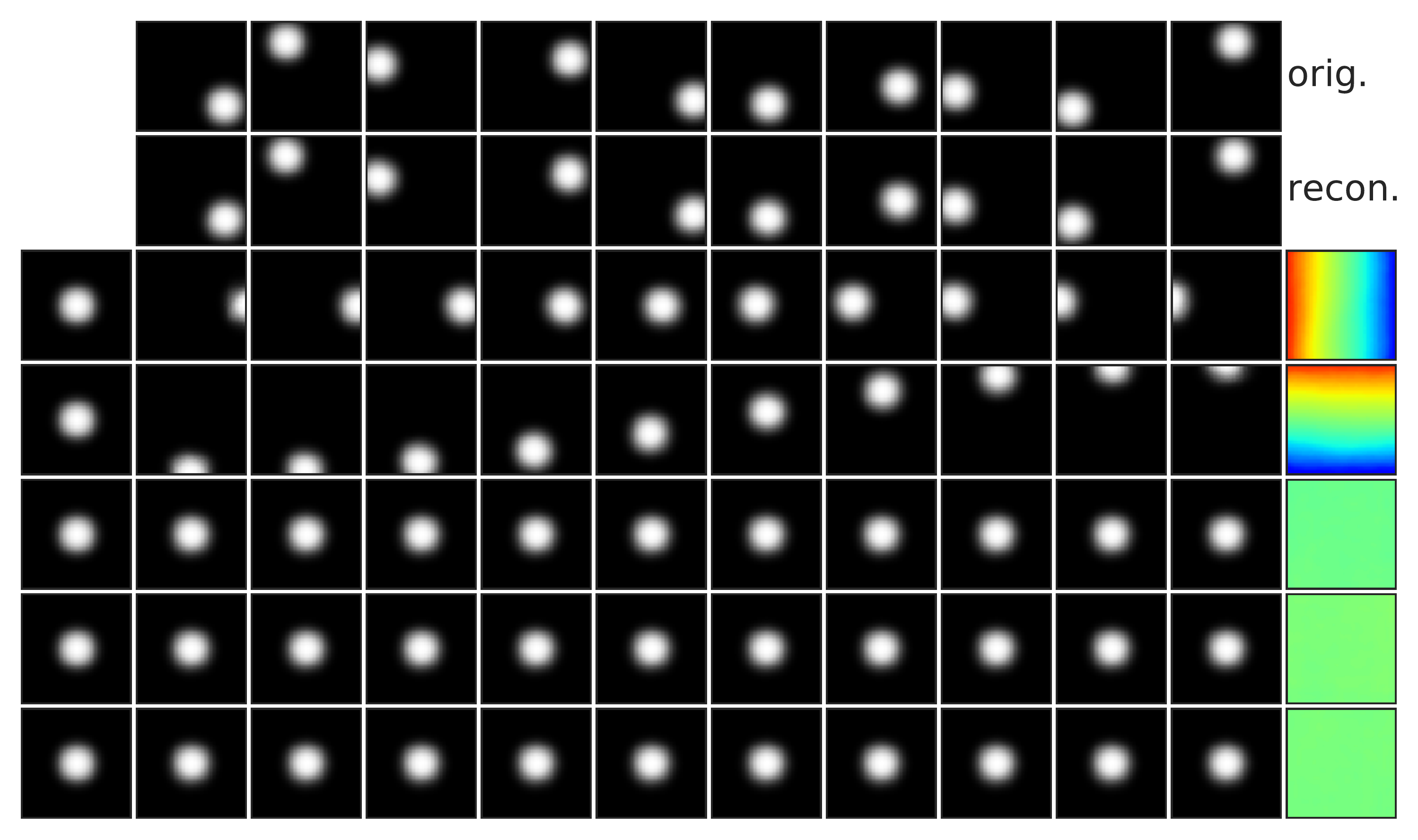}
    \label{fig:bestcvae}
    }
    \subfloat[Highest MSE ($621.44$)]{
    \includegraphics[width=.48\linewidth]{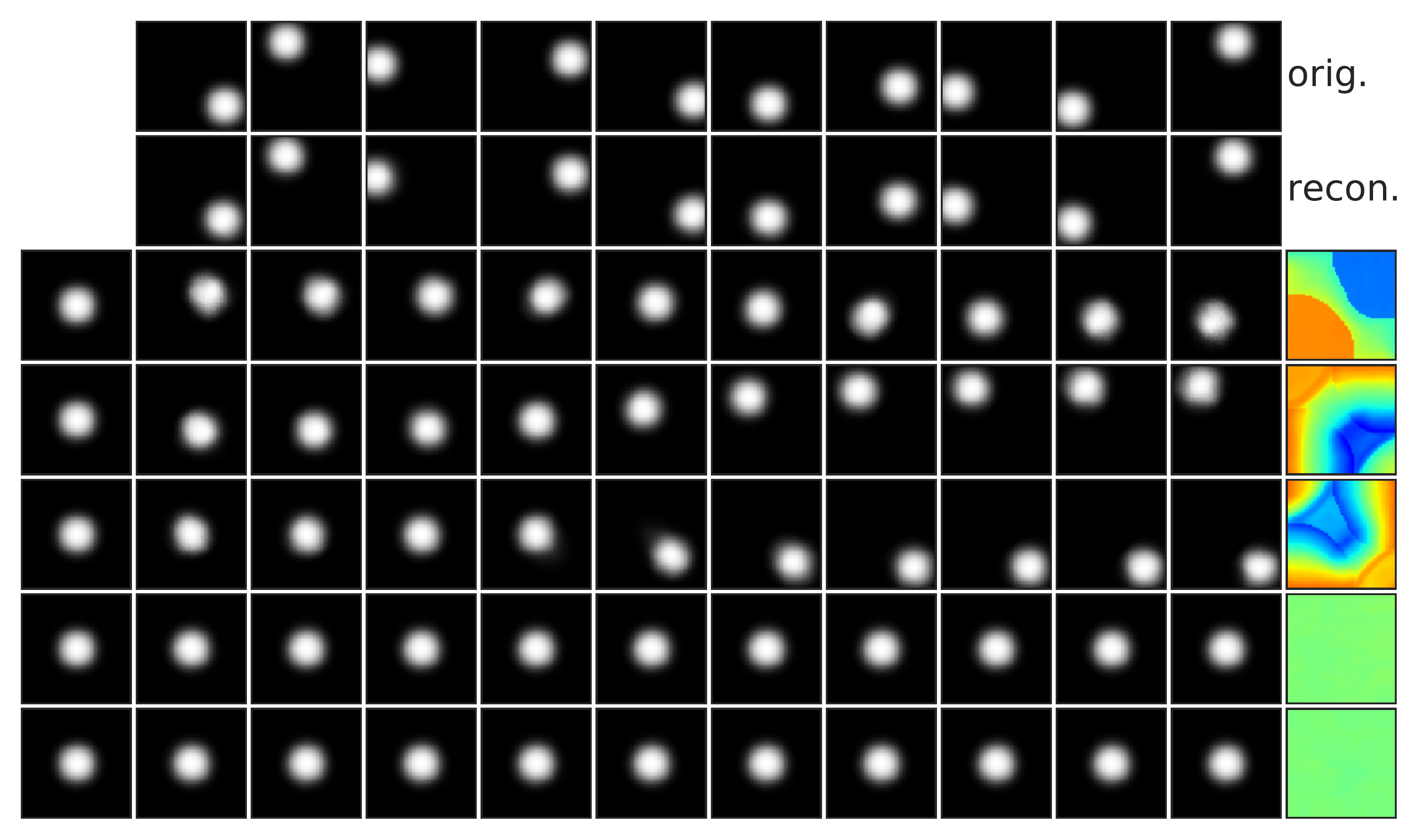}
    \label{fig:worsttcvae}
    }
    \caption{Visualization of representations learned by $\beta$-TCVAE with lowest and highest MSE out of 50 random seeds, measured for the task of learning the true Cartesian generative factors.}
    \label{fig:btcvaebaseline}
\end{figure}

We find that $\beta$-TCVAE for $\beta=40$ is able to learn a disentangled Cartesian representation without using unnecessary channels. 
As expected, the model is heavily dependent on random seeds \cite{LocatelloBLRGSB19}, as some learned representations were entangled and non-interpretable. 
Our specified MSE score used for the task of learning a Cartesian representation, was able to identify ``good'' and ``bad'' models.
In Figure \ref{fig:btcvaebaseline}, we visualize the latent traversals and positional heat maps of the models that achieved the lowest (Figure \ref{fig:bestcvae}) and highest (Figure \ref{fig:worsttcvae}) MSE out of 50 random weight initializations.
Evidently, the best model has clearly learnt the desired representation, in contrast to the worst model, where the representations are entangled and one redundant channel has been opened.

\begin{figure}[ht]
    \centering
    \subfloat[TCVAE: no labels, $\mathrm{MSE}{=}239.76$]{
    \includegraphics[width=.48\linewidth]{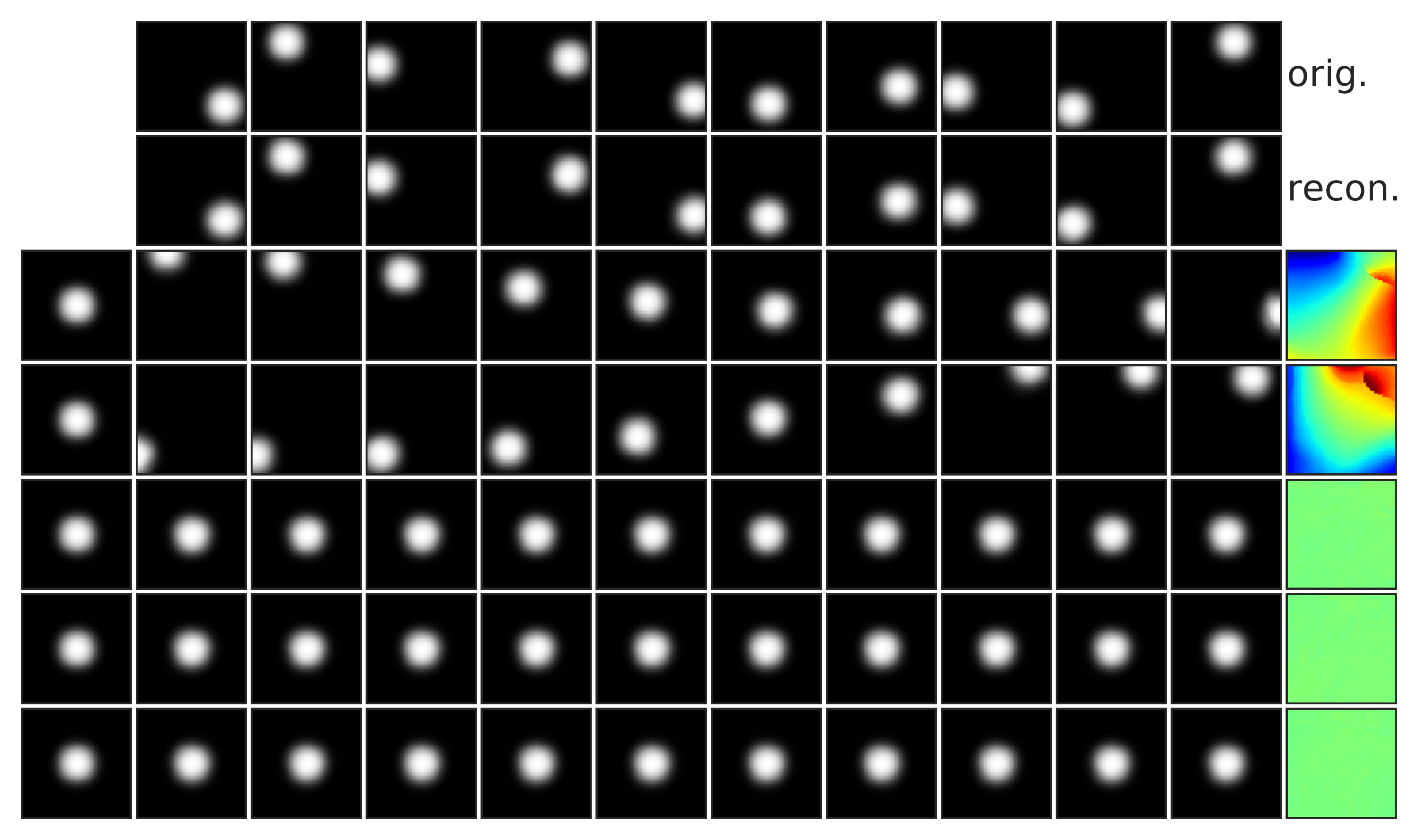}
    }
    \subfloat[WeLa-VAE: $p{=}2$, $\gamma{=}2000$, $\mathrm{MSE}{=}83.47$]{
    \includegraphics[width=.48\linewidth]{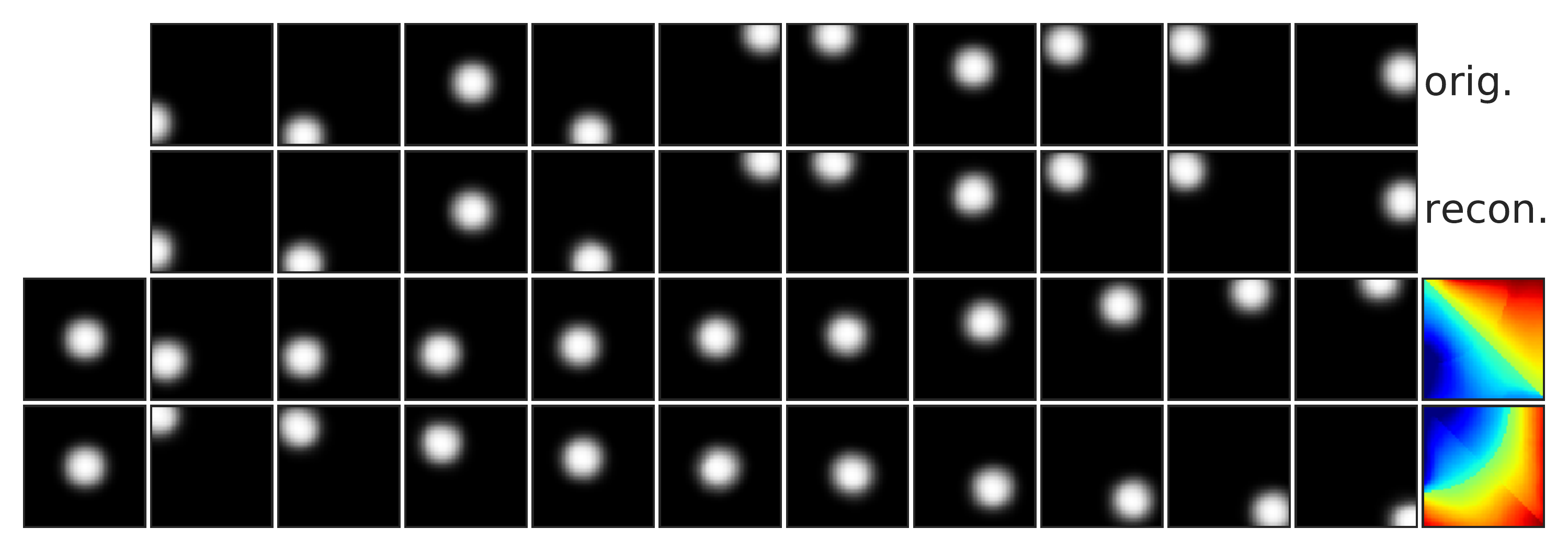}
    }\\
    \subfloat[WeLa-VAE: $p{=}3$, $\gamma{=}1500$, $\mathrm{MSE}{=}40.28$]{
    \includegraphics[width=.48\linewidth]{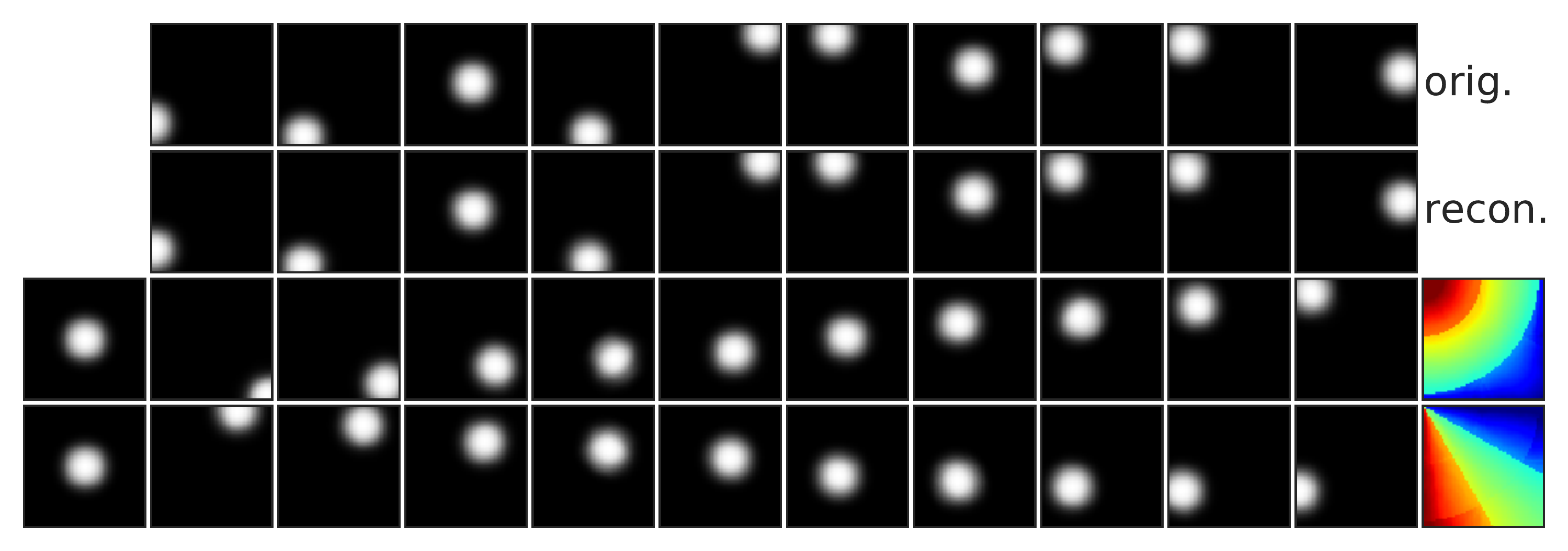}
    \label{fig:bestwelavae}
    } 
    \subfloat[WeLa-VAE: $p{=}4$, $\gamma{=}1000$, $\mathrm{MSE}{=}95.80$]{
    \includegraphics[width=.48\linewidth]{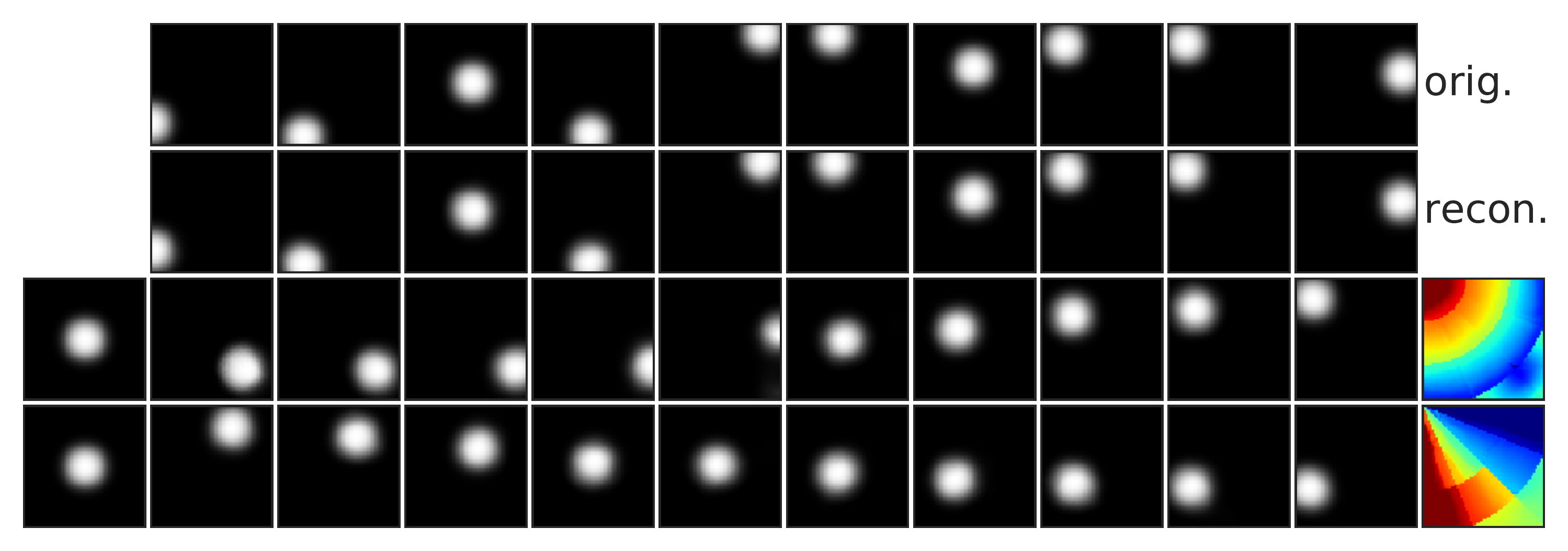}
    }\\
    \subfloat[WeLa-VAE: $p{=}5$, $\gamma{=}800$, $\mathrm{MSE}{=}176.04$]{
    \includegraphics[width=.48\linewidth]{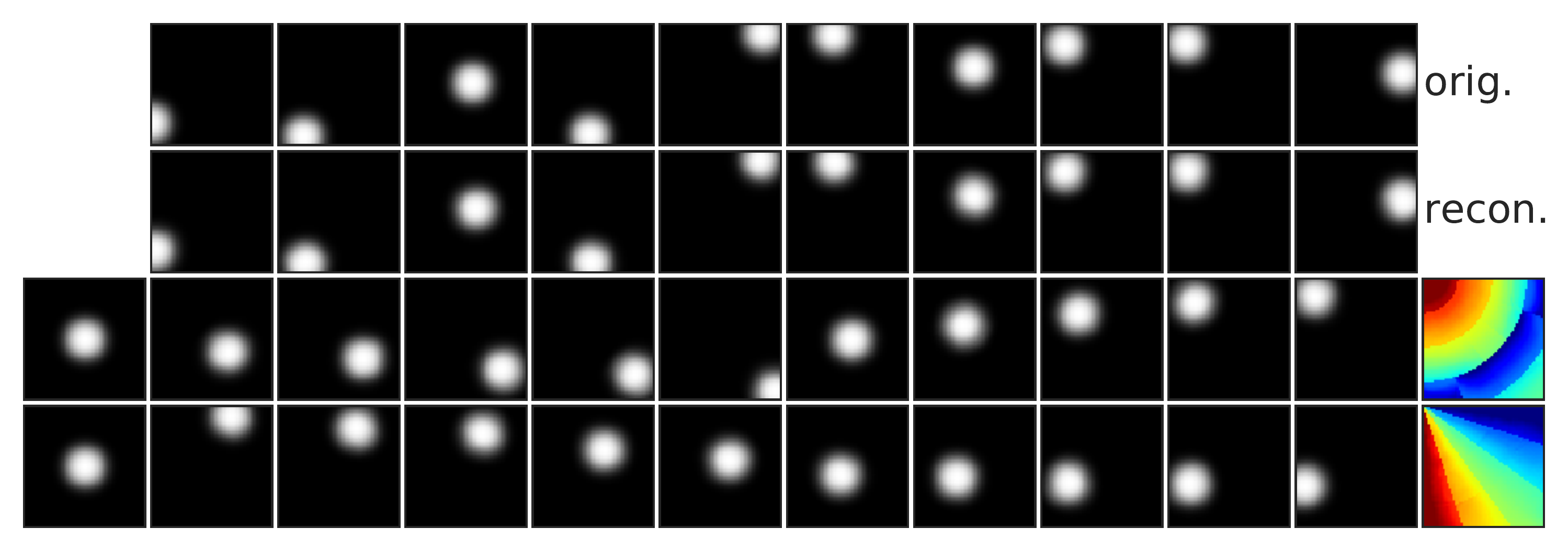}
    } 
    \subfloat[WeLa-VAE: $p{=}6$, $\gamma{=}750$, $\mathrm{MSE}{=}179.52$]{
    \includegraphics[width=.48\linewidth]{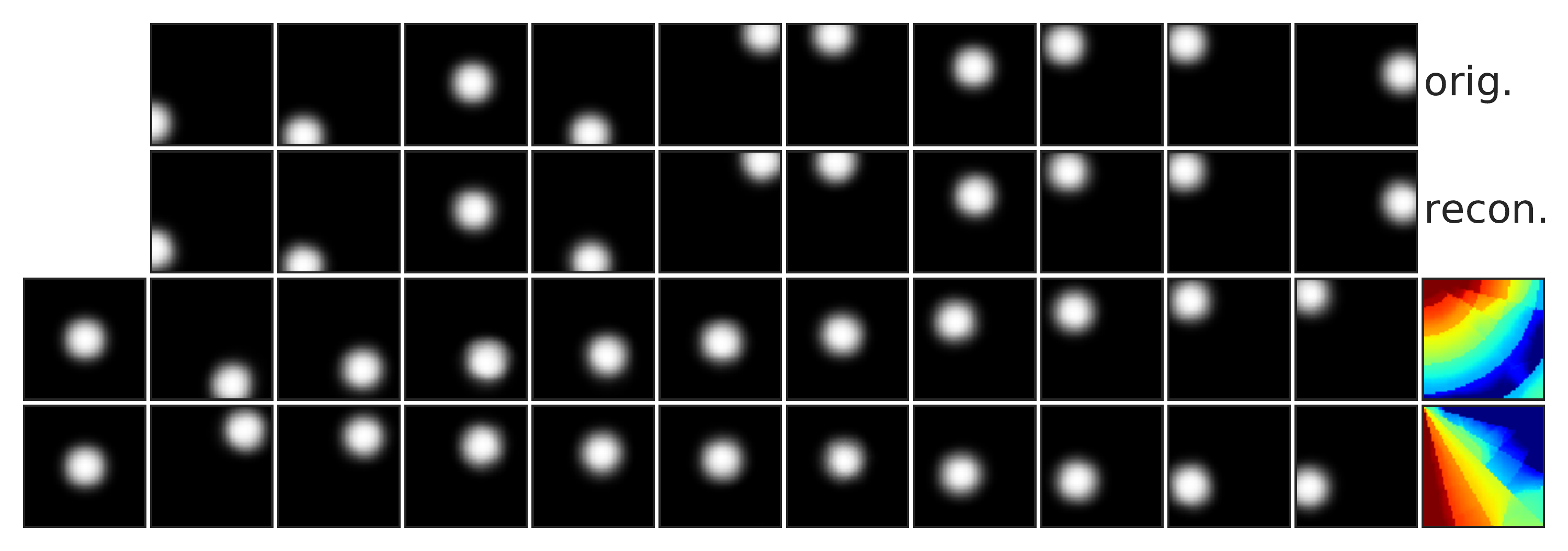}
    }\\
    \subfloat[WeLa-VAE: $p{=}7$, $\gamma{=}600$, $\mathrm{MSE}{=}190.44$]{
    \includegraphics[width=.48\linewidth]{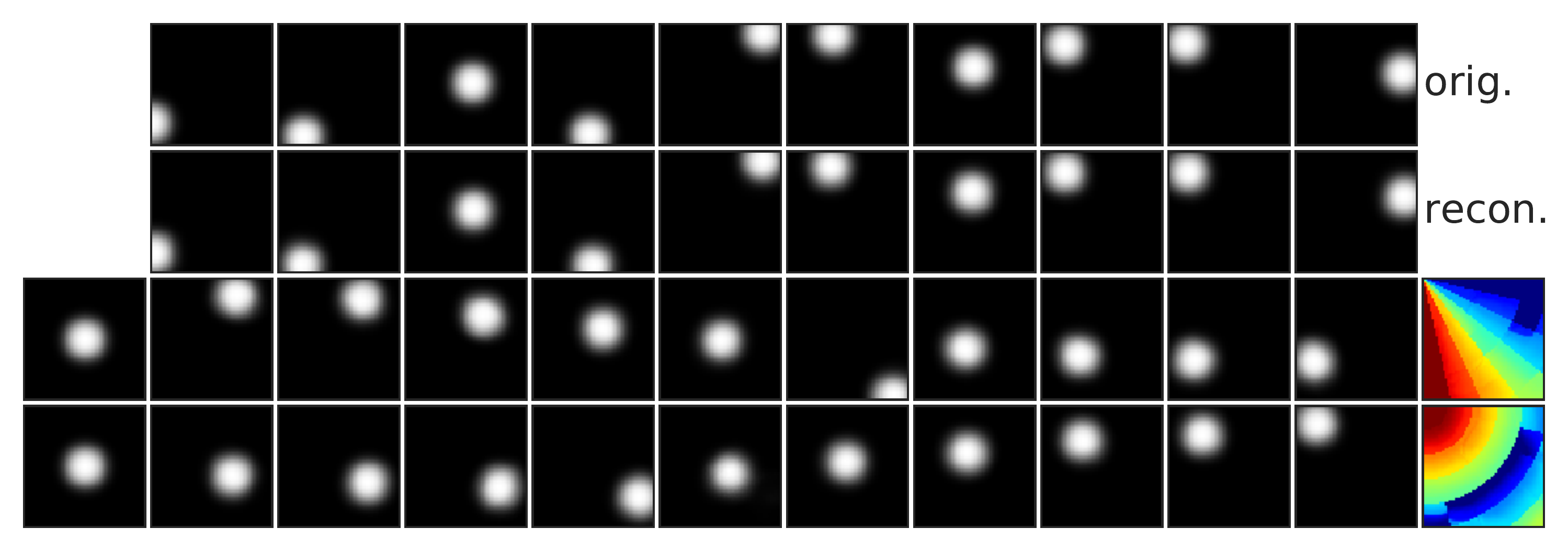}
    }
    \subfloat[WeLa-VAE: $p{=}8$, $\gamma{=}500$, $\mathrm{MSE}{=}160.74$]{
    \includegraphics[width=.48\linewidth]{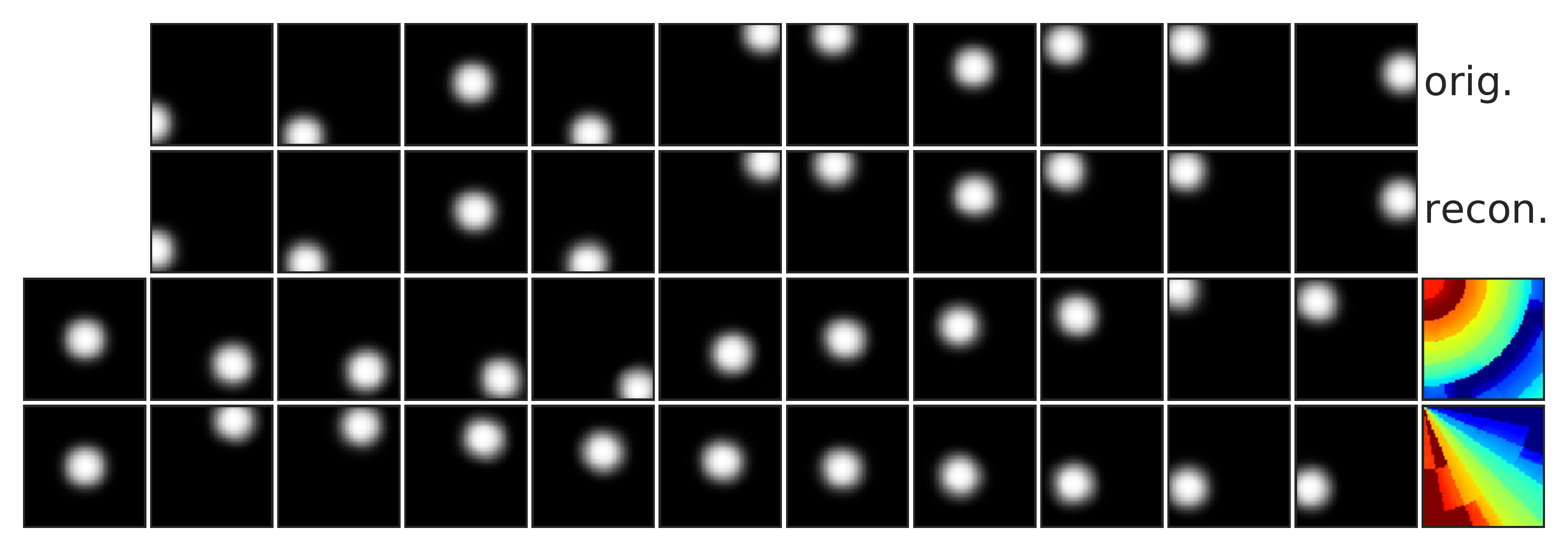}
    }
    \caption{WeLa-VAE trained on Blobs dataset with angle and distance labels of different dimensionality $p$. Each sub-figure visualizes the representations of the model with the lowest MSE out of 50 random seeds.}
    \label{fig:welavaeresults}
\end{figure}

For the task of learning a polar representation, we reused the same network and training configuration, as well as the $\beta$ value from the model of Figure \ref{fig:bestcvae} and trained WeLa-VAE for various label dimensionalities.
Table \ref{table:results} sums up the scores of all  models considered for this task, including the training accuracy of angle and distance of the best models.
The latent traversals and heat maps of the best models for TCVAE and WeLa-VAE are also shown in Figure \ref{fig:welavaeresults}.
WeLa-VAE was able to recover a clear polar representation for some label dimensionalities, suggesting that architectures and hyperparameters from unsupervised models can be transferred to WeLa-VAE, with the only requirement being the tuning of $\gamma$.
Moreover, label reconstruction error is low for all WeLa-VAE models, with training accuracy close to 100\% for both angle and distance, without compromising image reconstructions, which is of a similar quality to TCVAE.
As above, our evaluation metric identifies models whose representations exhibit the desired properties, i.e. representations with the best scores are the ones closer to being polar.
Indicatively, the model that learned the best representation uses a label dimensionality of $p=3$ (Figure \ref{fig:bestwelavae}), and is also the one achieving the total lowest MSE. 
On the other hand, no unsupervised TCVAE succeeded in effectively learning a polar representation.

However, MSEs indicate that WeLa-VAE is also heavily dependent on random seeds. 
Evidently, ``good'' models are scarce, with mean MSE being significantly greater than lowest MSE for all models.
Moreover, there is no evidence that higher label dimensionality $p$ had a positive effect on the representations, at least for the considered dataset.
On the contrary, better scores were obtained for low $p$ values, namely $p=2$ and $p=3$. 
Although $p=3$ yielded the best overall result, mean MSE and mean MSE of top 5 scores are significantly lower for $p=2$, compared to all $p$ values, meaning less sensitivity to random seeds and more frequent learning of ``good'' representations.

\begin{table}[ht]
    \centering
    \
    \caption{MSE scores on the task of learning a polar representation. For each model, we provide the hyperparameters and label dimensionality ($p$) used for training. We report the lowest, overall mean and mean of the best 10\% of scores, across 50 random seeds. We also report the training accuracy on angle ($\phi$) and distance ($d$) labels of the model with the lowest MSE.}
    \
    \begin{tabular}{ccccccccc}
        \toprule 
        \multirow{2}{*}{Model} & \multicolumn{3}{c}{Parameters} & \multicolumn{2}{c}{Accuracy} & \multicolumn{3}{c}{MSE} \\
        \cmidrule(l){2-4} \cmidrule(l){5-6} \cmidrule(l){7-9}\\
        {} &  $\beta$ & $\gamma$ & Label dim. & $\phi$ & $d$ & Lowest & Mean  & Mean: best 10\%  \\
        \midrule
        $\beta$-TCVAE & $40$ &-   & - & - & - & $239.76$   &  $435.68$ &  $286.14$\\
        \midrule
                 & & $2000$ & $p{=}2$ & $0.99$ & $1.00$ & $83.47$ & $\mathbf{255.28}$ & $\mathbf{94.30}$\\
                 & & $1500$ & $p{=}3$& $0.99$ & $1.00$ &  $\mathbf{40.28}$ & $824.88$ & $295.24$\\
                 & & $1000$ & $p{=}4$& $0.99$ & $1.00$ & $95.80$ & $751.90$ & $348.86$\\
        WeLa-VAE & $40$ & $800$ & $p{=}5$& $0.99$ & $1.00$ & $176.04$ & $797.66$ & $449.22$\\
                 & & $750$ & $p{=}6$& $0.99$ & $1.00$ & $179.52$ & $711.45$ & $315.62$\\
                 & & $600$ & $p{=}7$& $0.99$ & $1.00$ & $190.44$ & $686.62$ & $352.51$\\
                 & & $500$ & $p{=}8$& $0.99$ & $1.00$ & $160.74$ & $659.80$ & $275.18$\\
        \bottomrule
    \end{tabular}
    \label{table:results}
\end{table}

\section{Conclusion and future work} \label{section:conclusion}
We propose WeLa-VAE, a framework where weak labels are used to provide inductive biases towards disentanglement.
Our method is derived by assuming an extended generative model where labels $y$ and observations $x$ share the same latent variables.
The objective function is a modified variational lower bound of $\log \prior(x,y)$ with a weighted label reconstruction loss, coupled with a total correlation regularizer \cite{Kim2018, Chen2018} to enforce a factorized aggregated posterior.
We experiment with WeLa-VAE on a synthetic dataset consisting of images with two generative factors; Cartesian coordinates, for which a $\beta$-TCVAE model learns a disentangled, axis-aligned Cartesian representation.
Reusing the same architecture, optimization parameters and $\beta$ value of TCVAE, we provide WeLA-VAE with one-hot labels of varying dimension that indicate angle and distance.
We measure the performance of WeLa-VAE as the mean squared L2 error of the true generative factor values from an approximation obtained by a simple transformation of the learned representations.
This evaluation protocol discriminates representations with the desired qualities.
We find that WeLa-VAE successfully learns a disentangled polar representation with the cost of tuning only one extra hyperparameter.
Moreover, the best models used labels of low dimension, suggesting that WeLa-VAE does not need refined labels to perform well.
This shows that, weak supervision in the form of high-level labels provides the necessary inductive biases towards learning alternative and more interpretable disentangled representations.

An immediate application of WeLa-VAE is to aid interpretability, allowing for systematic and provable elicitation of alternative sets of interpretable and disentangled-enough representations.
An additional application of WeLa-VAE is related to the assumption that disentangled representations can lead to better performance on downstream tasks.
We expect that disentangled representations learned by WeLa-VAE can be useful for downstream tasks which relate to the given weak labels.
One limitation of WeLa-VAE is the requirement that latent channel size $K$ must be equal to the number of labels $m$, which is left as future work.
Moreover, the assumption of fully-observing the labels may be unrealistic in some scenarios. 
We conjecture that a variation of WeLa-VAE where labels are partially-observed would produce comparable results, which is currently under investigation.
Finally, an interesting direction for future work would be to remove the labels from the input layer of the encoder and let the decoder do all the work. 
Such a network could be trained in a dynamic setting where the labels become available incrementally and for a short amount of time, using learning-without-forgetting techniques  \cite{Li2018}.

\small
\bibliography{main}

\normalsize
\appendix
\section{Variational Lower Bound} \label{appendixA:elbo}
The generative model related to WeLa-VAE consists of $m+2$ random variables, $z$, $x$ and $y=(y_1, \ldots, y_m)$. First, the latent variable $z$ is sampled from a prior distribution $p(z)$. Then, both $x$ and $y$, which are conditionally independent with respect to $z$, are sampled from the posteriors $\prior(x|z)$ and $\prior(y|z) = \prod_{j=1}^m \prior(y_j | z)$, respectively. We are interested in obtaining a variational lower bound on the log-likelihood of the joint distribution $\prior(x,y)$.

Applying the chain rule yields
\begin{align*}
    \log{\prior(x,y)} = \log{\frac{\prior(y,x,z)}{\prior(z|x,y)}} &= \log{\frac{\prior(y|z) \cdot \prior(x|z) \cdot p(z) }{\prior(z|x,y)}}.
\end{align*} 
Then, we introduce the auxiliary distribution $\post(z|x,y)$ which approximates the intractable $\prior(z|x,y)$ on the right-hand side, and take the expectation over all $z$ sampled from $\post(z|x,y)$, which gives
\begin{align*}
    \EE_{z \sim \post} [\log{\prior(x,y)}] &= \EE_{z \sim \post} \left[ \log{\frac{\prior(y|z) \cdot \prior(x|z) \cdot p(z)}{\prior(z|x,y)}} \cdot \frac{\post(z|x,y)}{\post(z|x,y)} \right]\\
    &= \EE_{z \sim \post} \left[ \log{\frac{\post(z|x,y)}{\prior(z|x,y)}} \right]  + \EE_{z \sim \post} \left[ \log{\frac{\prior(y|z) \cdot \prior(x|z) \cdot p(z)}{\post(z|x,y)}} \right]\\
    &= \dkl{\post(z|x,y)}{\prior(z|x,y)} + \EE_{z \sim \post} \left[ \log{\frac{\prior(y|z) \cdot \prior(x|z) \cdot p(z)}{\post(z|x,y)}} \right].
\end{align*}
Since KL divergence is non-negative and $\prior(x,y)$ is constant with respect to $z$,
\begin{align*}
    \log{\prior(x,y)} &\geq \EE_{z \sim \post} \left[ \log{\frac{\prior(y|z) \cdot \prior(x|z) \cdot p(z)}{\post(z|x,y)}} \right] \\
    &= \EE_{z \sim \post}[\log{\prior(x|z)}] + \EE_{z \sim \post}[\log{\prior(y|z)}] + \EE_{z \sim \post} \left[ -\log{\frac{\post(z|x,y)}{p(z)}} \right]\\
    &= \EE_{z \sim \post}[\log{\prior(x|z)}] + \EE_{z \sim \post}[\log{\prior(y|z)}] -\dkl{\post(z|x,y)}{p(z)}.
\end{align*}

\end{document}